\documentclass[conference]{IEEEtran}
\usepackage{cite}
\usepackage{amsmath,amssymb,amsfonts}
\usepackage{graphicx}
\usepackage{textcomp}
\usepackage{cite}
\usepackage{amsmath,amssymb,amsfonts}
\usepackage{graphicx}
\usepackage{graphicx}%
\usepackage{multirow}%
\usepackage{amsmath,amssymb,amsfonts}%
\usepackage{amsthm}%
\usepackage{mathrsfs}%
\usepackage{textcomp}%
\usepackage{manyfoot}%
\usepackage{booktabs}%
\usepackage{algorithmicx}%
\usepackage{algpseudocode}%
\usepackage{listings}%
\usepackage{lastpage,fancyhdr,graphicx}
\usepackage{epstopdf}
\usepackage{comment}
\usepackage{amsmath,amssymb}

\usepackage{changepage}

\usepackage{textcomp,marvosym}
\usepackage{hyperref}       
\usepackage{url}            
\usepackage{booktabs}       
\usepackage{amsfonts}       
\usepackage{nicefrac}       
\usepackage{booktabs}
\usepackage{lipsum}
\usepackage{fancyhdr}       
\usepackage{graphicx}       
\usepackage{amsmath}
\usepackage{float}
\usepackage{subcaption}
\usepackage{wrapfig}
\usepackage{graphicx}
\usepackage{booktabs}

\usepackage{adjustbox}
\usepackage{multirow}    
\usepackage{booktabs}    

\usepackage{textcomp}
\usepackage{xcolor}
\def\BibTeX{{\rm B\kern-.05em{\sc i\kern-.025em b}\kern-.08em
    T\kern-.1667em\lower.7ex\hbox{E}\kern-.125emX}}
\begin{document}

\title{Can Vision-Language Models Understand Construction Workers? An Exploratory Study}

\author{
\IEEEauthorblockN{Hieu Bui}
\IEEEauthorblockA{
\textit{Department of Electrical and Computer Engineering} \\
\textit{Villanova University} \\
Villanova, PA, USA \\
hbui01@villanova.edu
}
\and
\IEEEauthorblockN{Nathaniel E. Chodosh}
\IEEEauthorblockA{
\textit{Department of Computing Sciences} \\
\textit{Villanova University} \\
Villanova, PA, USA \\
nathaniel.chodosh@villanova.edu
}
\and
\IEEEauthorblockN{Arash Tavakoli}
\IEEEauthorblockA{
\textit{Department of Civil and Environmental Engineering} \\
\textit{Villanova University} \\
Villanova, PA, USA \\
arash.tavakoli@villanova.edu
}
}

\maketitle

\begin{abstract}
As robotics become increasingly integrated into construction workflows, their ability to interpret and respond to human behavior will be essential for enabling safe and effective collaboration. Vision-Language Models (VLMs) have emerged as a promising tool for visual understanding tasks and offer the potential to recognize human behaviors without extensive domain-specific training. This capability makes them particularly appealing in the construction domain, where labeled data is scarce and monitoring worker actions and emotional states is critical for safety and productivity. In this study, we evaluate the performance of three leading VLMs-GPT-4o, Florence 2, and LLaVa-1.5-in detecting construction worker actions and emotions from static site images. Using a curated dataset of 1,000 images annotated across ten action and ten emotion categories, we assess each model’s outputs through standardized inference pipelines and multiple evaluation metrics. GPT-4o consistently achieved the highest scores across both tasks, with an average F1-score of 0.756 and accuracy of 0.799 in action recognition, and an F1-score of 0.712 and accuracy of 0.773 in emotion recognition. Florence 2 performed moderately, with F1-scores of 0.497 (action) and 0.414 (emotion), while LLaVa-1.5 showed the lowest overall performance (F1-scores of 0.466 for action and 0.461 for emotion). Confusion matrix analyses revealed that all models struggled to distinguish semantically close categories-such as “Collaborating in teams” versus “Communicating with supervisors,” or “Focused” versus “Determined”-highlighting limitations in current VLMs when applied to visually nuanced, domain-specific tasks. While the results indicate that general-purpose VLMs can offer a baseline capability for human behavior recognition in construction environments, further improvements-such as domain adaptation, temporal modeling, or multimodal sensing-may be needed for real-world reliability. This study provides an initial benchmark and practical insights for deploying human-aware AI systems in complex, safety-critical settings.
\end{abstract}


\section{Introduction}

The construction industry is transforming rapidly with the introduction of robotic systems-from drones and autonomous vehicles to quadruped robots-designed to improve job site efficiency, safety, and monitoring \cite{marinelli2022human,shayesteh2021investigating,shayesteh2022enhanced,nassar2024human}. These agents must navigate complex, unstructured environments while collaborating with human workers. However, for effective collaboration, robots must go beyond obstacle avoidance or object recognition to understand human behavior \cite{zhang2023human,li2020influence,de2020future,shayesteh2022enhanced,shayesteh2021investigating}.

This understanding encompasses not only worker actions but also their physiological and emotional states, whether during tasks or in isolation \cite{alsulami2023impact,chong2022impact,hwang2018measuring,langdon2018construction}. Much like physical activities, cognitive states such as fatigue or anxiety critically influence safety, decision-making, and site coordination \cite{chong2022impact,albeaino2023psychophysiological}. Consequently, recent research has increasingly focused on automating state detection using wearable sensors \cite{wang2024monitoring,ahn2019wearable,hwang2017wristband,cheng2022measuring} and computer vision \cite{liu2021applications,li2024review,roberts2020vision} to support real-time monitoring and adaptive decision-making.

While traditional computer vision models rely on large labeled datasets and lack generalizability \cite{zhang2024vision,laurenccon2024matters}, Vision-Language Models (VLMs) offer a robust alternative. For instance, task-specific architectures like YOLO \cite{jiang2022review} often require extensive retraining to adapt to new contexts. Conversely, VLMs such as GPT-4o integrate visual and linguistic processing to interpret complex scenes through natural language reasoning \cite{zhang2024vision}. Although these models excel in general multimodal tasks, their ability to understand human behavior in safety-critical construction environments remains largely unexplored \cite{alsulami2023impact, braga2024robotic}.

This study is an explanatory approach to investigating whether general-purpose VLMs, without fine-tuning for the construction domain, can accurately detect both the actions and emotional states of construction workers from still images, compared to labels generated by a human annotator. Our work paves the way for integrating VLMs into real-time monitoring systems that support human-robot collaboration in complex and safety-critical environments. By examining the performance of three state-of-the-art models on a curated dataset of labeled construction site images, we aim to evaluate their ability to serve as a foundation for human-aware robotic systems in real-world job sites. The rest of the paper is organized as follows: we begin by reviewing related work on worker state monitoring and vision-language modeling. We then describe our dataset, labeling process, and experimental pipeline, followed by an evaluation of model performance. Finally, we discuss the implications of our findings, limitations, and directions for future research.

\section{Background Information}
Automating site monitoring is a growing priority in the architecture, engineering, and construction (AEC) industry \cite{musarat2024automated,qureshi2022factors}. Given the dynamic and hazardous nature of construction environments, non-intrusive tracking of worker behavior, equipment, and site progress is essential for enhancing safety, productivity, and real-time decision-making \cite{gatti2014physiological,aryal2017monitoring,rao2022real}. In line with the automation goals, robotics adoption in construction is gaining momentum due to their potential to enhance productivity, efficiency, and worker safety by taking on hazardous tasks \cite{liu2022human}. As the interaction level between the robot and the human increases, the communication of needs, states, and expectations between the two entities become more necessary \cite{aaltonen2018refining}. In line with this expectation, studies have started to propose various methods to understand and predict workers’ states in real time. 

Worker state can be broadly defined as the physical, cognitive, and emotional condition of a worker at any given time, encompassing aspects such as fatigue, stress, attentiveness, workload, and overall well-being \cite{anwer2021evaluation,cheng2022measuring,hwang2018measuring}. Accurately assessing these states is critical in high-risk construction environments, where lapses in attention or physical exhaustion can result in accidents or decreased performance \cite{liu2022human,nassar2024human,aaltonen2018refining,shayesteh2021investigating}. In this regard, prior studies collectively show that worker's physiological measures, facial expression, actions, and contextual metrics can be used as an input for predicting various states such as cognitive workload, performance metrics (e.g., attention level), or stress when collaborating with or without a robotic agent \cite{liu2022human,nassar2024human,aaltonen2018refining,shayesteh2021investigating,yu2019automatic,gao2022immersive}. On the physiological side, studies have demonstrated the utility of cardiovascular metrics, eye tracking, and brain activity measures for recognizing worker states such as fatigue \cite{aryal2017monitoring,anwer2023identification,li2020identification,lee2017wearable}, stress \cite{jebelli2018eeg,jebelli2019application,umer2022simultaneous,patching2014investigation}, and cognitive load \cite{hasanzadeh2017measuring,chen2018measuring,han2020eye}. While these physiological indicators are valuable, they often require the use of additional wearable sensors, which may interfere with workers' comfort or freedom of movement on-site.

To address these limitations, researchers are increasingly exploring vision-based approaches that rely solely on video data to predict worker states without intrusive sensors \cite{yu2019automatic,li2020identification}. These studies demonstrate the potential of computer vision to assess both physical actions and cognitive conditions. For instance, Yu et al. monitored physical fatigue using a vision-based 3D motion capture system and biomechanical modeling; they demonstrated that joint-level fatigue can be assessed automatically using only RGB video, eliminating the need for wearables \cite{yu2019automatic}. Similarly, Li et al. investigated mental fatigue in equipment operators using eye-tracking, showing that supervised learning models (specifically SVM) can classify fatigue levels with up to 85\% accuracy \cite{li2020identification}. Similar approaches were taken for understanding worker actions using computer vision \cite{roberts2020vision,sherafat2020automated,ishioka2020single}. For instance, Roberts et al. developed a deep learning–based framework that uses 2D pose estimation and RGB video footage to automatically classify worker activities during tasks such as bricklaying and plastering, achieving over 78\% accuracy in activity recognition. Their work demonstrates how worker posture, derived from vision-based pose tracking, can serve as a key input for activity classification and ergonomic assessment on construction sites \cite{roberts2020vision}.

Despite these advances, the deployment of computer vision on construction sites still faces several limitations, including environmental variability, occlusion, and the high cost of labeled data. Among these, the issue of acquiring labeled data is particularly challenging \cite{zhang2024vision}. Traditional computer vision models require large volumes of task-specific annotations-an impractical requirement given the diversity and nuance of construction worker activities and conditions \cite{zhang2024vision}. The integration of vision and language understanding-enabled by Vision-Language Models (VLMs)-offers a promising alternative. Large Language Models (LLMs) are deep learning models trained on vast amounts of textual data to perform tasks such as language generation, translation, summarization, and question answering \cite{chang2024survey}. These models, such as GPT-4 \cite{achiam2023gpt}, and BERT \cite{devlin2019bert}, have demonstrated strong generalization capabilities and the ability to perform reasoning and inference through natural language prompts.

Vision Large Language Models (VLMs) extend traditional capabilities by incorporating visual inputs alongside text \cite{zhang2024vision,zhou2022learning}. Trained on large-scale image-text pairs, these multimodal models align visual features with language representations to jointly interpret both data types \cite{zhang2024vision}. This enables them to perform diverse tasks-such as visual question answering (VQA), image captioning, and multimodal dialogue-without explicit re-training \cite{gao2025application,zhang2024vision,fan2024assisting,cao2024visdiahalbench}. By leveraging natural language prompts and contextual reasoning, VLMs offer a flexible, generalizable approach for interpreting complex human behaviors on dynamic construction sites, potentially reducing the dependence on extensive labeled datasets \cite{zhang2024vision}.

Various VLMs have been proposed in recent years. Initial VLM work by \cite{Radford2021CLIP} used image-text pairs to align the feature spaces of an image and text encoder. They showed surprising success on zero-shot image classification by simply comparing image features to the text encoding of potential labels. Subsequent work developed algorithms for generating better prompts and other improvements in inference (see \cite{Menon2023VisualClassification,Pratt2023Platypus,Roth2023Waffling}). However, new chat-based models such as GPT-4o (\cite{zhang2024vision}), Florence 2 (\cite{xiao2024florence}), and LLaVa-1.5 (\cite{liu2023visualinstructiontuning}) integrate image, audio, and text modalities in a unified interface, enabling them to answer questions about visual content. When evaluated through the chat interface rather than through feature-space classification, \cite{Cooper2025Rethinking} and \cite{Zhang2024VisuallyGrounded} found that these models generally perform worse than previous VLMs. However, the simplicity of the chat interface makes it likely to be used in real-world construction monitoring where machine-learning expertise is not available. Therefore, we aim to study the performance of these models when used in their most user-friendly form.


\section{Research Questions}
Current VLMs are trained on broad, general-purpose datasets and lack optimization for task-specific applications like construction site monitoring. This study assesses the generalizability of these models for behavior recognition in high-risk, high-variability construction environments. Specifically, we explore whether general-purpose VLMs can detect both physical actions and emotional states from static images without fine-tuning, benchmarking their performance against human annotators. We evaluate three state-of-the-art models-GPT-4o, Florence 2, and LLaVa-1.5-on a curated dataset labeled with diverse worker behaviors, ranging from actions (e.g., operating machinery) to emotions (e.g., focused, tired, anxious). Using standard classification metrics (precision, recall, F1-score, per-class accuracy) and confusion matrix analyses, we examine each model’s ability to interpret complex visual and contextual cues relevant to worker states.
Specifically, we investigate the following research questions:

\begin{itemize} 
\item Can general-purpose VLMs accurately detect construction-related worker actions from static images without task-specific training? 
\item Can these models recognize emotional and cognitive states-such as being focused, tired, or cooperative-based solely on visual information? 
\item How do GPT-4o, Florence 2, and LLaVa-1.5 compare in their performance across these recognition tasks? 
\item What patterns of misclassification emerge, and what do they reveal about model limitations in domain-specific behavioral interpretation? \end{itemize}

Rather than aiming for definitive performance benchmarks, this exploratory work seeks to probe the current capabilities of VLMs in a real-world, safety-critical domain. The findings are intended to inform future research on adapting multimodal AI for human-centered monitoring and robotic collaboration in construction and similar applied settings.

\section{Methods}

\begin{figure}
        \centering
        \includegraphics[width=\columnwidth]{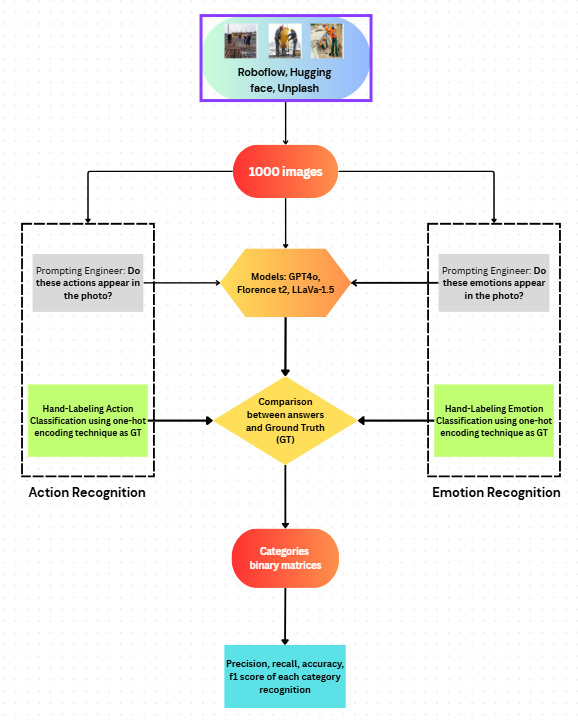}
        \caption{Framework of the study}
        \label{fig:framework}
\end{figure}

We designed a multi-step experimental pipeline that includes data collection, manual annotation, model selection, inference, and quantitative evaluation using standard classification metrics (Fig. \ref{fig:framework}). Below we detail out each part of the pipeline. 

\subsection{Data Collection}
As stated, we focus specifically on workers actions as a behavioral metric and emotional situations as a state metric. We considered the following actions: \textit{operating heavy machinery, lifting and carrying materials, measuring and marking surfaces, mixing cement or concrete, using hand tools such as hammers and drills, inspecting completed work, collaborating in teams, communicating with supervisors, following safety protocols, and taking breaks.} For the states herein referred to as emotions, we considered \textit{being focused, determined, tired, alert, satisfied, anxious, proud, frustrated, cooperative, and relieved.}

We begin the process by collecting 1000 images of construction scenes, using a mix of online open-source image repositories (e.g. Unsplash \cite{unsplash_construction}) as well as using Roboflow construction images dataset \cite{testttt2025outputmerge, amitslonimski2022ppetection, apd2024ppetection, qatrunnadas2023ppe, deulsaha2023soda, testcasque2024ppe, akfa2023safetyrd, kwonsoonbin2025detectworker}. We note that we intentionally used images from prior available datasets as our goal was not to create a new dataset, but rather to build on those commonly used in the field. 

\subsection{Data Labeling}

\begin{figure*}[t]
  \centering

  \includegraphics[width=\textwidth]{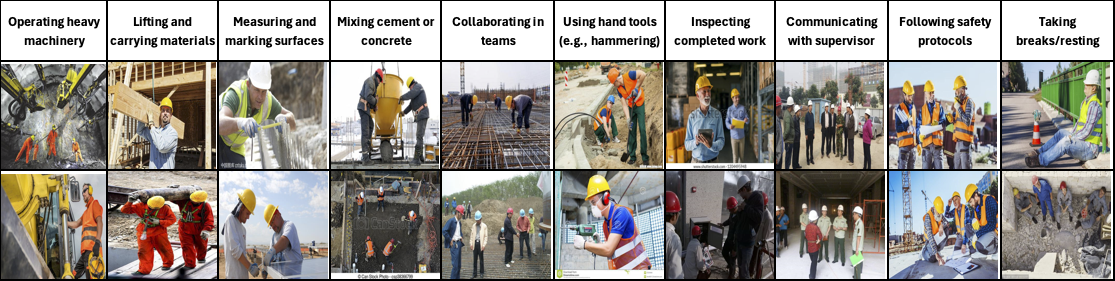}\par
  \vspace{0.6\baselineskip}

  \includegraphics[width=\textwidth]{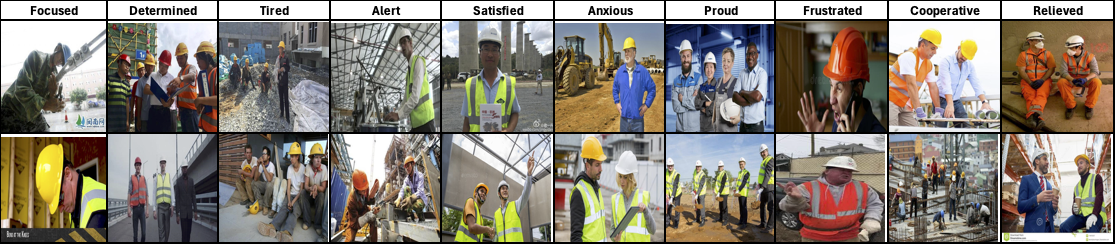}

  \caption{Examples of labeled construction site images for the action and emotion categories used in this study.}
  \label{fig:sample_images}
\end{figure*}

While the data may have had specific labels, we are specifically focusing on actions and emotions in the images, hence we needed to relabel all the images. For each category, we labeled data using the one-hot coding technique. If an image has a specific action or emotion, it will be labeled as ``1'', ``0'' if it does not. Subsequently, we processed the results as a binary matrix of 1000 x 10, where each row is an array containing the data of each photo. The labeled categories as well as a sample of the images in each category can be seen on Fig. \ref{fig:sample_images}. 

\subsection{Model Selection}
For this study, three specific VLMs were considered: GPT4o, LLaVa-1.5  \cite{liu2023visualinstructiontuning}, and Microsoft Florence 2 models \cite{xiao2024florence-}. They are relatively strong models with outstanding performance on object detection \cite{achiam2023gpt}.

\subsection{Inference Pipeline}

After constructing the labeled dataset, we developed a standardized inference pipeline to evaluate the performance of each model. Florence 2 and LLaVa-1.5 were retrieved from the Hugging Face Transformers library \cite{huggingface_transformers}, while GPT-4o was accessed through the OpenAI API. Each model was adapted to perform binary classification on the ten action categories and ten emotion categories.

We follow the ``Direct Generation'' inference strategy from \cite{Zhang2024VisuallyGrounded} for zero-shot classification. For each model, images were processed through customized inference scripts that prompted the model to predict the presence or absence of each action or emotion by asking \textit{Does the action/emotion of XXXX appear in the photo?}. The model outputs were collected in structured JSON files, where each image was associated with a list of ``Yes'' or ``No'' responses for all categories. These outputs were subsequently converted into binary matrices of size 1000 × 10 for both actions and emotions, where ``Yes'' responses were encoded as 1 and ``No'' responses as 0.

The same conversion was applied to the manually labeled ground truth data, resulting in two reference matrices for comparison. Model outputs were then quantitatively evaluated against the ground truth using standard metrics, including Precision, Recall, F1-score, and Accuracy, calculated independently for each category. All inference scripts and data processing pipelines were standardized across models to ensure fair comparison. The full source code and dataset are available at \cite{hieubui_construction}.

\subsection{Evaluation}
After collecting and organizing the evaluation matrices for each model, we begin by performing a thorough comparison using key performance metrics, such as Precision, Recall, F1-Score, and Accuracy. Each model's output is evaluated against the ground-truth data, which consists of hand-labeled matrices representing the true classifications. More specifically, precision measures the proportion of correct positive predictions made by the model out of all predicted positive cases. In contrast, recall measures the proportion of actual positive cases correctly identified by the model. Both metrics are calculated for each class (i.e., action or emotion labels), providing a detailed insight into how well the model handles false positives and false negatives.

Mathematically, Precision and Recall for each class are computed as follows:

\begin{align*}
\text{Precision} &= \frac{TP}{TP + FP} \\
\text{Recall} &= \frac{TP}{TP + FN} \\
\end{align*}

Where:
\begin{itemize}
    \item TP (True Positives): Correctly identified positive instances.
    \item FP (False Positives): Incorrectly predicted as positive.
    \item FN (False Negatives): Incorrectly predicted as negative.
\end{itemize}

The F1-Score is a harmonic mean of Precision and Recall, providing a single metric to balance both the false positives and false negatives in the model's predictions. This is particularly useful when the class distribution is imbalanced. The F1-score for each class is computed as:

\begin{align*}
\text{F1-Score} &= \frac{2 \cdot (\text{Precision} \cdot \text{Recall})}{\text{Precision} + \text{Recall}}
\end{align*}

This score allows us to measure the model's overall ability to correctly identify the positive instances while minimizing errors, making it a robust evaluation metric. In addition to Precision, Recall, and F1-Score, Accuracy serves as a baseline metric, representing the proportion of correct predictions across all classes. While useful, accuracy might not always provide a comprehensive picture, especially in cases with imbalanced class distributions. By considering these metrics together, we gain a comprehensive understanding of each model's strengths and weaknesses. For example, a model with high accuracy but low recall might be overfitting to certain classes, whereas a model with a high F1-score likely strikes a better balance between precision and recall.

\section{Results}

\subsection{Qualitative Analysis}

\begin{figure}[H]
    \centering

    \begin{subfigure}[c]{0.48\columnwidth}  
        \centering
        \includegraphics[width=\columnwidth]{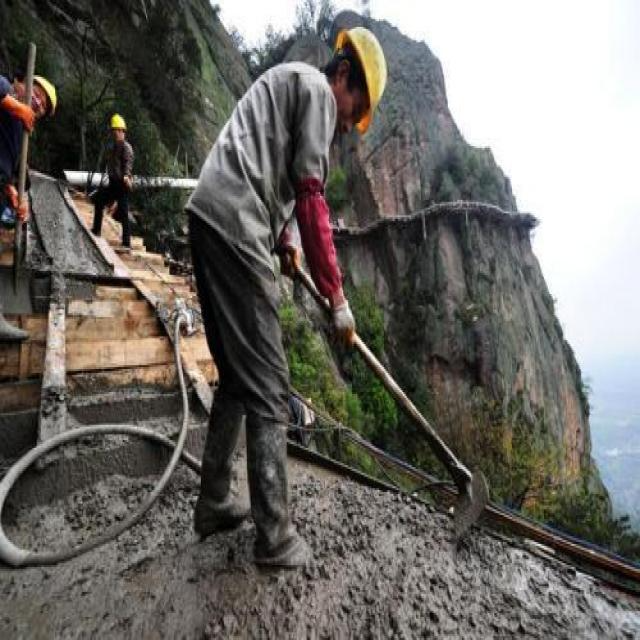}
        \caption{Sample construction image.}
        \label{fig:qualitative_image}
    \end{subfigure}
    \hfill
    \begin{subfigure}[c]{0.48\columnwidth}  
        \centering
        \scriptsize
        \begin{adjustbox}{max width=\columnwidth}
        \begin{tabular}{llccc}
        \toprule
        \textbf{Category} & \textbf{Label} & \textbf{GPT-4o} & \textbf{Florence 2} & \textbf{LLaVa-1.5} \\
        \midrule
        \multirow{5}{*}{Action}
        & Heavy machinery    & 0 & 1 & 1 \\
        & Carrying materials & 0 & 1 & 1 \\
        & Marking surfaces   & 0 & 1 & 1 \\
        & Mixing concrete    & 0 & 1 & 0 \\
        & Team collaboration & 1 & 1 & 1 \\
        \midrule
        \multirow{5}{*}{Emotion}
        & Focused            & 1 & 1 & 1 \\
        & Determined         & 1 & 0 & 1 \\
        & Tired              & 1 & 0 & 1 \\
        & Alert              & 0 & 0 & 1 \\
        & Satisfied          & 1 & 0 & 1 \\
        \bottomrule
        \end{tabular}
        \end{adjustbox}
        \caption{Model predictions for selected actions and emotions.}
        \label{tab:qualitative_predictions}
    \end{subfigure}

    \caption{Qualitative illustration of model predictions: (Left) sample construction image; (Right) corresponding predictions from GPT-4o, Florence 2, and LLaVa-1.5.}
    \label{fig:qualitative_combined}
\end{figure}

To better illustrate model behavior, we conducted a qualitative analysis on a representative construction site image. Figure~\ref{fig:qualitative_combined} displays the image alongside model predictions for ten action categories and ten emotion categories, compared to the ground truth (GT) labels.

For action recognition, all three models correctly identified collaborative tasks such as ``Team collaboration,'' ``Using hand tools,'' and ``Inspecting completed work.'' However, Florence 2 and LLaVa-1.5 both incorrectly predicted the presence of ``Operating heavy machinery,'' ``Carrying materials,'' and ``Marking surfaces,'' where no such activities were present, suggesting a tendency to over-detect active construction behaviors. GPT-4o showed more conservative predictions, correctly avoiding these false positives but missing the true instance of ``Mixing concrete,'' resulting in a false negative.

For emotion recognition, all models accurately detected ``Focused,'' reflecting the strong visual cues associated with concentration. However, both GPT-4o and LLaVa-1.5 incorrectly predicted the presence of ``Tired'' despite the ground truth label being negative, suggesting difficulty in differentiating subtle affective states based on static imagery. LLaVa-1.5, in particular, tended to overpredict emotions such as ``Alert,'' ``Anxious,'' and ``Frustrated,'' demonstrating less discrimination between distinct emotional/cognitive states. Florence 2 appeared more conservative but missed several correct positive labels, including ``Determined'' and ``Satisfied.''

\subsection{Quantitative Analysis}

\begin{figure}[H]
    \centering
    \resizebox{0.7\columnwidth}{!}{%
        \includegraphics{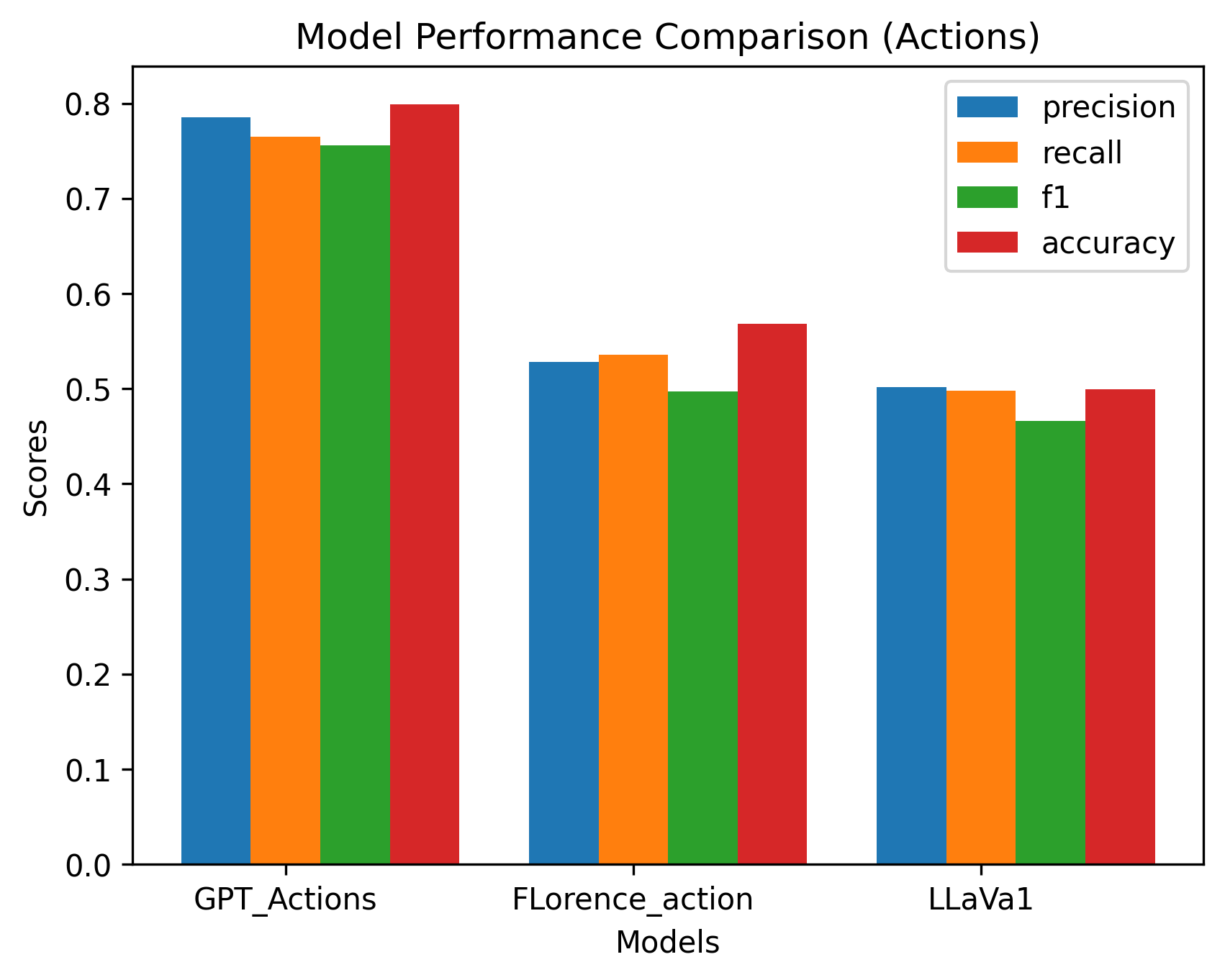}
    }
    \caption{Actions-evaluation Performance}
    \label{fig:act1}
\end{figure}

Figure~\ref{fig:act1} presents an overall performance comparison across the key metrics of precision, recall, F1-score, and accuracy for the action recognition task. As shown, GPT-4o achieved the highest overall scores, with a precision of 0.7854, recall of 0.7648, F1-score of 0.7560, and accuracy of 0.7988. Florence 2 showed moderate performance, with a precision of 0.5279, recall of 0.5356, F1-score of 0.4971, and accuracy of 0.5681. LLaVa-1.5 performed the weakest, attaining a precision of 0.5017, recall of 0.4980, F1-score of 0.4664, and accuracy of 0.4996. These results confirm that GPT-4o maintained a strong balance between precision and recall, while the other models, particularly LLaVa-1.5, struggled with consistent classification across the action dataset.

Table~\ref{tab:per-class-accuracy} provides a detailed breakdown of each model’s accuracy across specific construction-related action categories. GPT-4o demonstrated robust performance across nearly all categories, with especially high accuracy in “Following safety protocols” (0.915), “Mixing cement or concrete” (0.884), and “Taking breaks/resting” (0.880). These results suggest GPT-4o is particularly effective at recognizing actions with distinctive visual cues or contextual markers.

Interestingly, Florence 2, despite its overall lower performance, achieved competitive results in several classes-for instance, it closely followed GPT-4o in identifying “Taking breaks/resting” (0.834 vs. 0.880) and showed solid accuracy in “Mixing cement or concrete” (0.830). This indicates that Florence 2 may still be effective in capturing certain well-structured physical activities, albeit with lower consistency overall.

LLaVa-1.5, while slightly outperforming Florence 2 in isolated tasks such as “Using hand tools” (0.522 vs. 0.486), generally trailed in most categories. Its limited accuracy in classes like “Measuring and marking surfaces” (0.459) and “Collaborating in teams” (0.490) highlights its weaker generalization and potential confusion across visually similar tasks.

\begin{table}[htbp]
\centering
\caption{Per-Class Accuracy by Model and Action Category}
\label{tab:per-class-accuracy}
\resizebox{\linewidth}{!}{%
\begin{tabular}{lccc}
\toprule
\textbf{Action Category} & \textbf{GPT4o} & \textbf{Florence 2} & \textbf{LLaVa-1.5} \\
\midrule
Operating heavy machinery           & 0.762 & 0.533 & 0.510 \\
Lifting and carrying materials      & 0.786 & 0.594 & 0.523 \\
Measuring and marking surfaces      & 0.718 & 0.632 & 0.459 \\
Mixing cement or concrete           & 0.884 & 0.830 & 0.476 \\
Collaborating in teams              & 0.833 & 0.641 & 0.490 \\
Using hand tools (e.g., hammering, drilling) & 0.628 & 0.486 & 0.522 \\
Inspecting completed work           & 0.753 & 0.223 & 0.491 \\
Communicating with supervisors      & 0.829 & 0.649 & 0.506 \\
Following safety protocols          & 0.915 & 0.259 & 0.505 \\
Taking breaks/resting               & 0.880 & 0.834 & 0.514 \\
\bottomrule
\end{tabular}%
}
\end{table}

\begin{table}[htbp]
\centering
\caption{Action Evaluation Metrics for GPT-4o, Florence 2, and LLaVa-1.5}
\label{tab:action_metrics}
\resizebox{\linewidth}{!}{%
\begin{tabular}{lcccc}
\toprule
\textbf{Model} & \textbf{Precision} & \textbf{Recall} & \textbf{F1 Score} & \textbf{Accuracy} \\
\midrule
GPT-4o        & 0.7854 & 0.7648 & 0.7560 & 0.7988 \\
Florence 2   & 0.5279 & 0.5356 & 0.4971 & 0.5681 \\
LLaVa-1.5     & 0.5017 & 0.4980 & 0.4664 & 0.4996 \\
\bottomrule
\end{tabular}%
}
\end{table}

\begin{figure}[H]
    \centering
    \includegraphics[width=0.7\columnwidth]{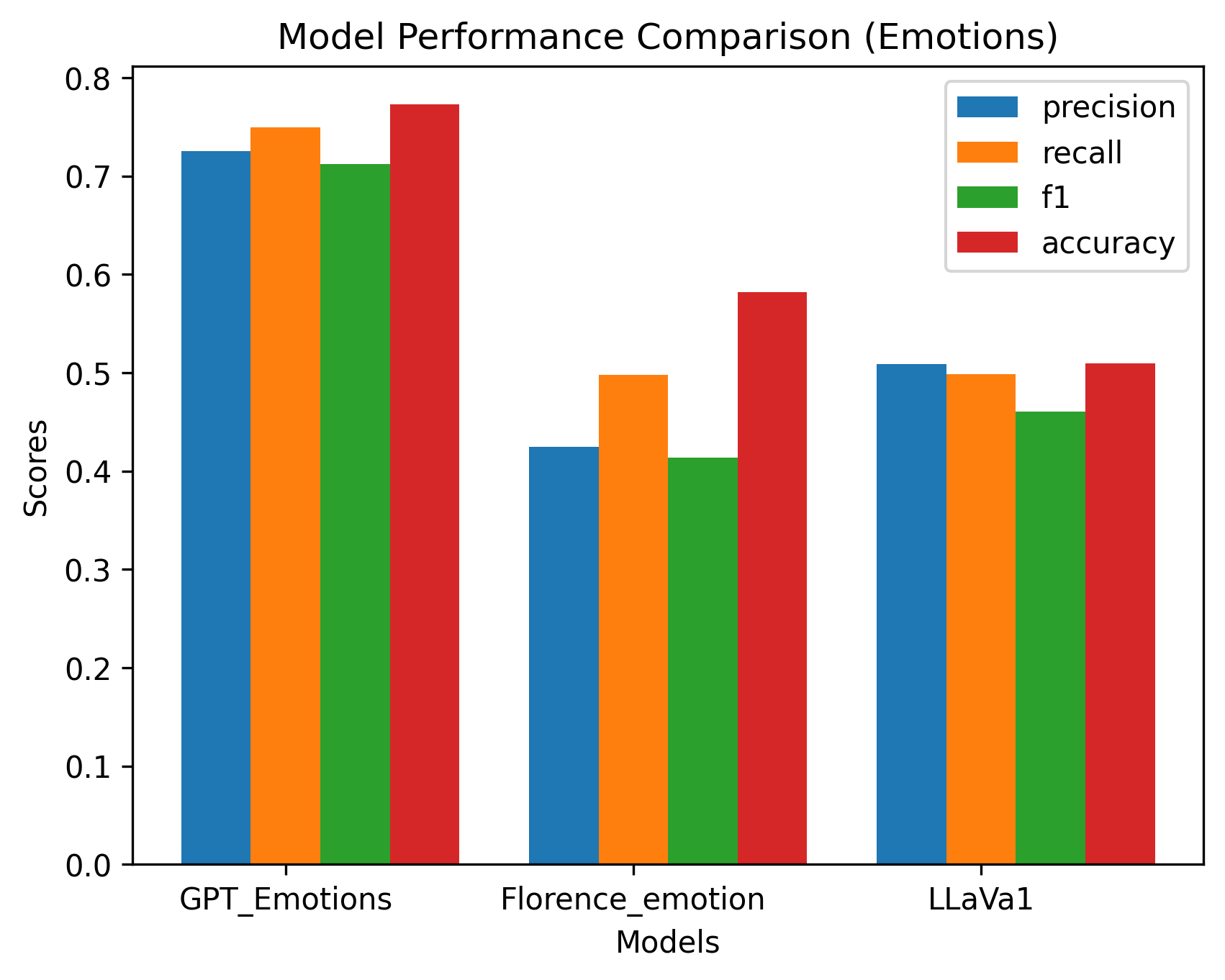}
    \caption{Emotions-evaluation Performance} 
    \label{fig:emot1}
\end{figure}

We also evaluated the models on their ability to recognize worker emotional and cognitive states from construction site images. The evaluation was conducted on a dataset of 1000 images annotated across ten emotion categories. Figure~\ref{fig:emot1} presents the overall performance of the three models using precision, recall, F1-score, and accuracy as evaluation metrics.

GPT-4o outperformed the other models, achieving the highest scores across all four metrics: precision of 0.7254, recall of 0.7496, F1-score of 0.7124, and accuracy of 0.7729. These results indicate strong and consistent performance in identifying emotional states from visual input. Florence 2 showed moderate performance, with a precision of 0.4247, recall of 0.4977, F1-score of 0.4140, and accuracy of 0.5822, suggesting difficulty in distinguishing between subtle emotional expressions. LLaVa-1.5, while slightly outperforming Florence 2 in terms of precision (0.5088), lagged behind in recall (0.4985) and accuracy (0.5098), pointing to inconsistent emotional classification.

Table~\ref{tab:per-class-accuracy-emotions} provides a more detailed view of per-class accuracy across the ten emotion categories. GPT-4o achieved the highest accuracy in nearly all categories, including standout performance in ``Frustrated'' (0.967), ``Tired'' (0.943), and ``Relieved'' (0.886). These results suggest that GPT-4o is especially effective at identifying emotions that are more visually distinct or contextually grounded.

Florence 2, while generally trailing GPT-4o, delivered competitive results in several categories-most notably ``Relieved'' (0.829), ``Proud'' (0.656), and ``Cooperative'' (0.637)-demonstrating its potential to detect certain positive or contextually consistent emotional states. LLaVa-1.5 displayed relatively flat performance across categories, with most class accuracies hovering around 0.50. Its best result, “Frustrated” (0.531), still fell noticeably short of GPT-4o’s accuracy, suggesting a need for targeted improvements in emotional disambiguation.

\begin{table}[htbp]
\centering
\caption{Per-Class Accuracy by Model and Emotion Category}
\resizebox{\linewidth}{!}{%
\begin{tabular}{lccc}
\toprule
\textbf{Emotion Category} & \textbf{GPT4o} & \textbf{Florence 2} & \textbf{LLaVa-1.5} \\
\midrule
Focused       & 0.698 & 0.403 & 0.509 \\
Determined    & 0.691 & 0.400 & 0.514 \\
Tired         & 0.943 & 0.690 & 0.512 \\
Alert         & 0.470 & 0.480 & 0.494 \\
Satisfied     & 0.786 & 0.654 & 0.519 \\
Anxious       & 0.851 & 0.625 & 0.514 \\
Proud         & 0.764 & 0.656 & 0.509 \\
Frustrated    & 0.967 & 0.448 & 0.531 \\
Cooperative   & 0.673 & 0.637 & 0.488 \\
Relieved      & 0.886 & 0.829 & 0.508 \\
\bottomrule
\end{tabular}%
}
\label{tab:per-class-accuracy-emotions}
\end{table}

\begin{table}[htbp]
\centering
\caption{Emotion Evaluation Metrics}
\label{tab:emotion-metrics}
\resizebox{\linewidth}{!}{%
\begin{tabular}{lcccc}
\toprule
\textbf{Model} & \textbf{Precision} & \textbf{Recall} & \textbf{F1 Score} & \textbf{Accuracy} \\
\midrule
GPT-4o       & 0.7254 & 0.7496 & 0.7124 & 0.7729 \\
Florence 2  & 0.4247 & 0.4977 & 0.4140 & 0.5822 \\
LLaVa-1.5    & 0.5088 & 0.4985 & 0.4606 & 0.5098 \\
\bottomrule
\end{tabular}%
}
\end{table}

\subsection{Confusion Matrix Analysis}
\begin{figure*}[!t]
\centering
\setlength{\tabcolsep}{6pt} 
\renewcommand{\arraystretch}{1.0}

\begin{tabular}{cc}
\subfloat[GPT-4o (Action)\label{fig:GPTa}]
{\includegraphics[width=0.47\textwidth]{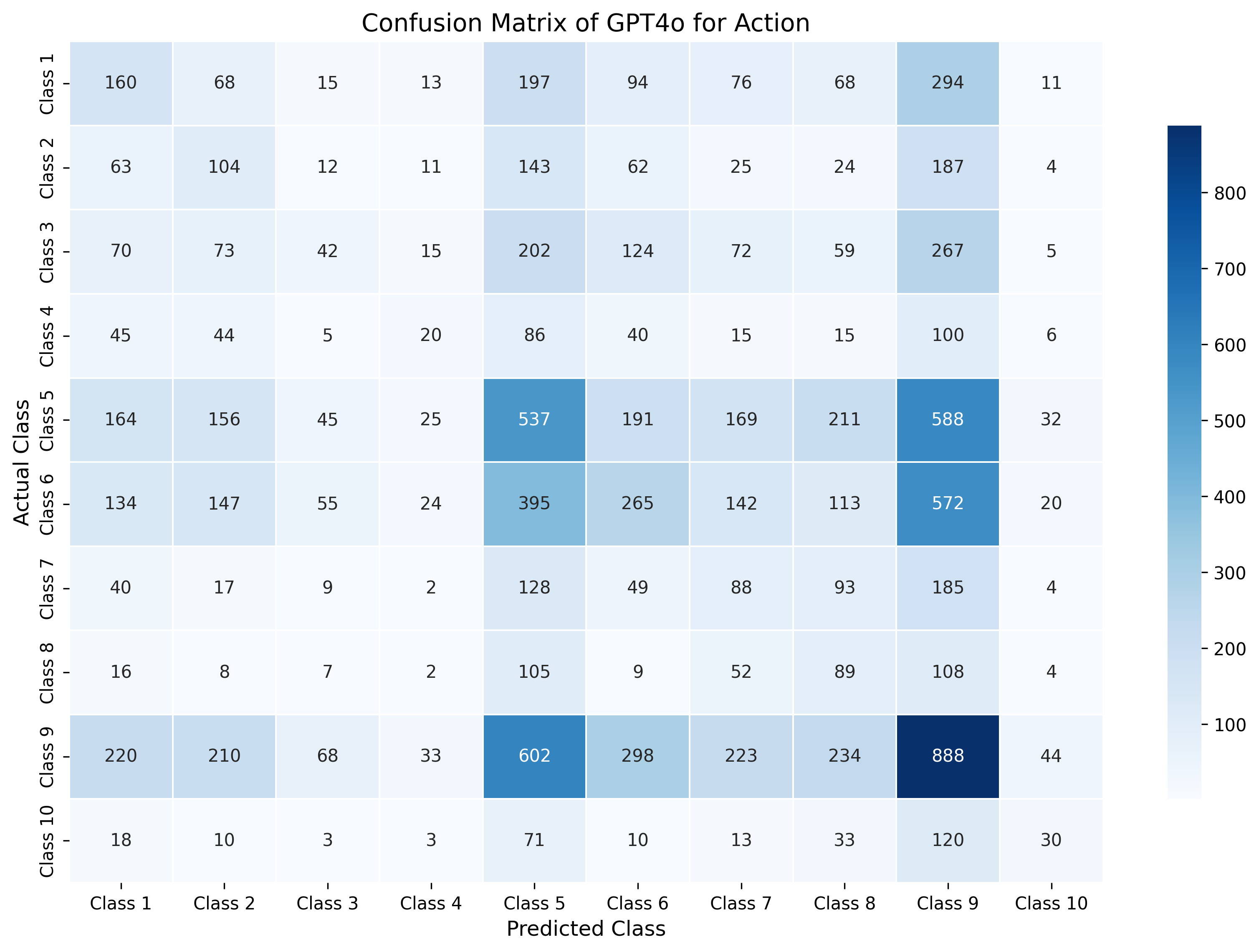}} &
\subfloat[GPT-4o (Emotion)\label{fig:GPTe}]
{\includegraphics[width=0.47\textwidth]{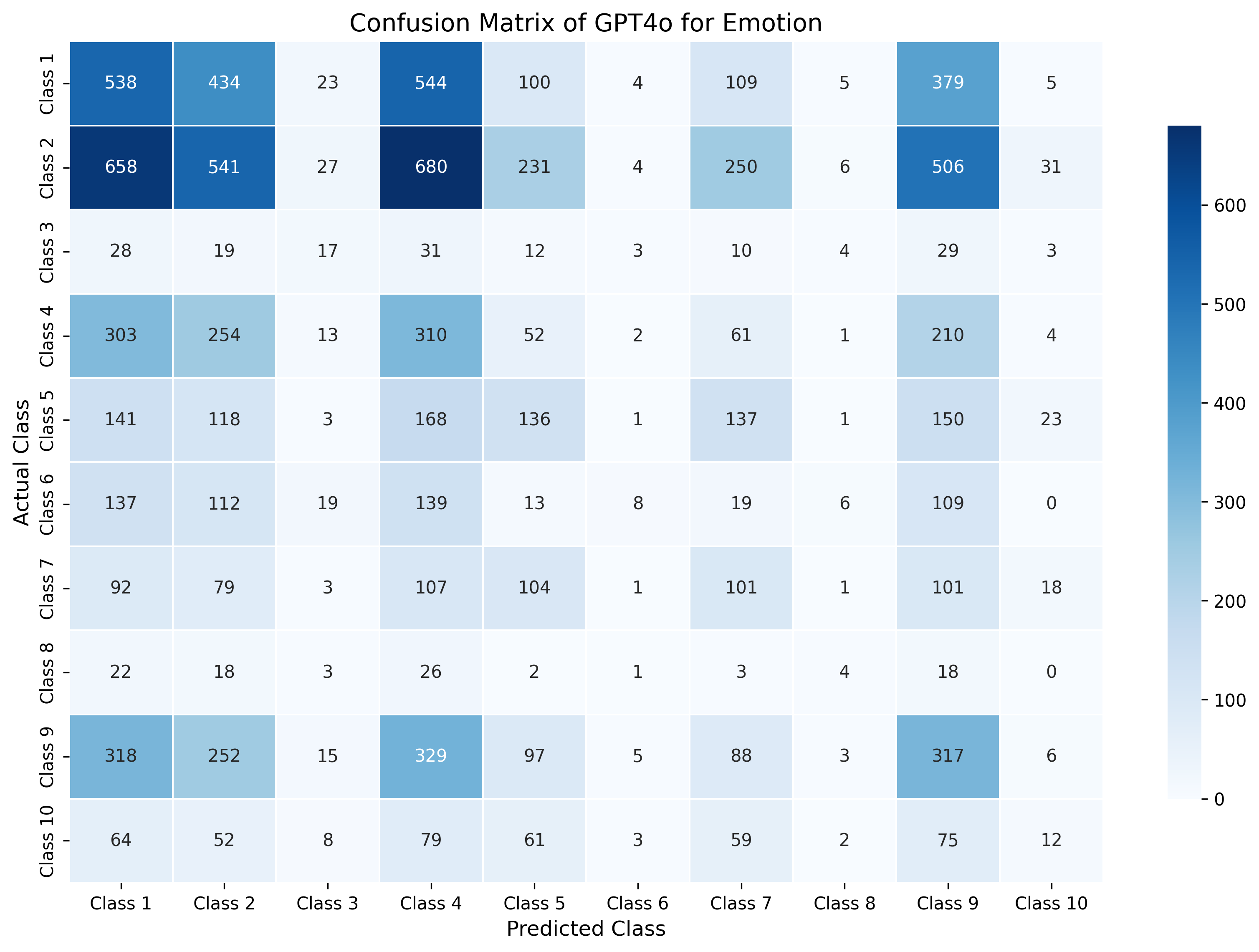}} \\[6pt]

\subfloat[Florence 2 (Action)\label{fig:F2a}]
{\includegraphics[width=0.47\textwidth]{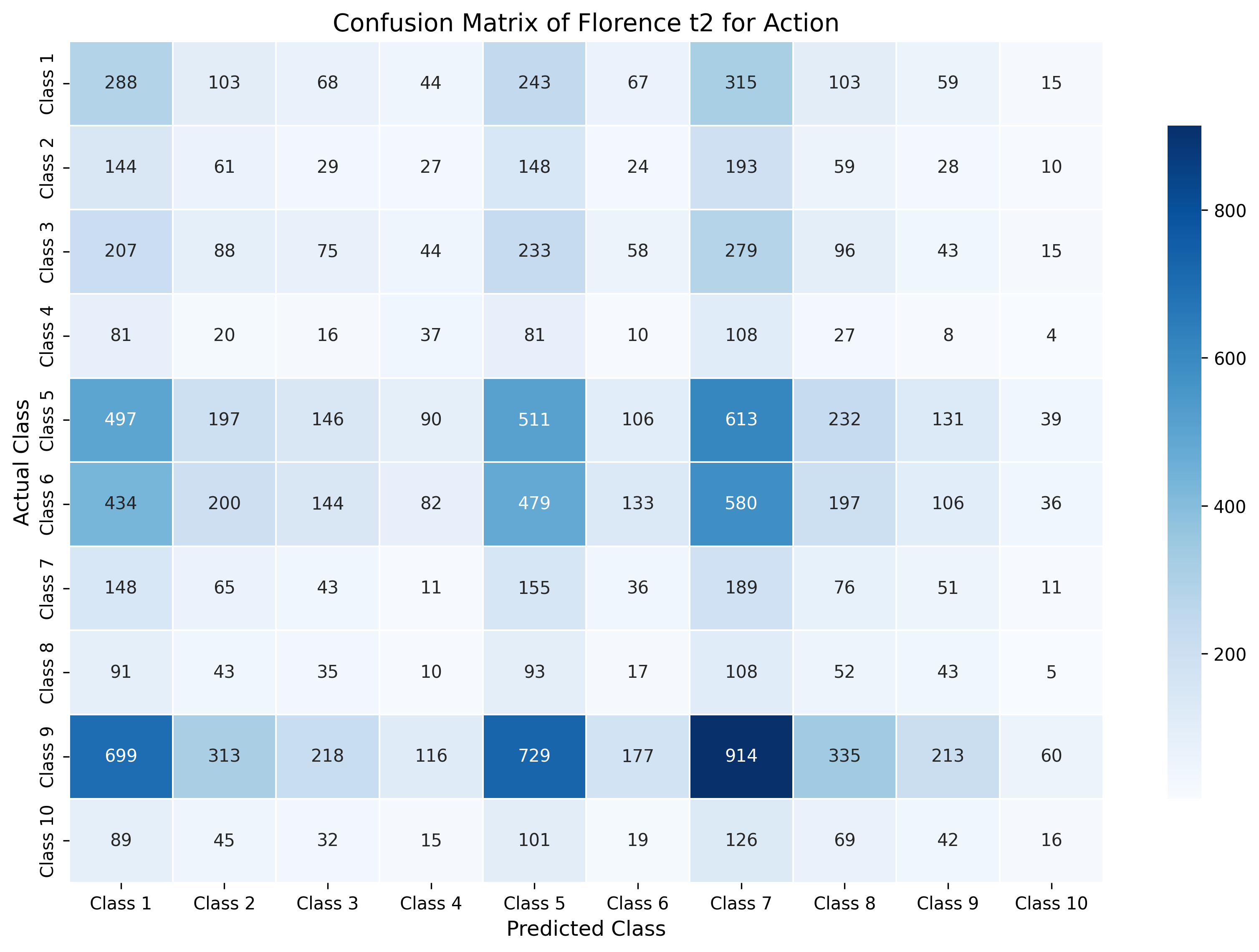}} &
\subfloat[Florence 2 (Emotion)\label{fig:F2e}]
{\includegraphics[width=0.47\textwidth]{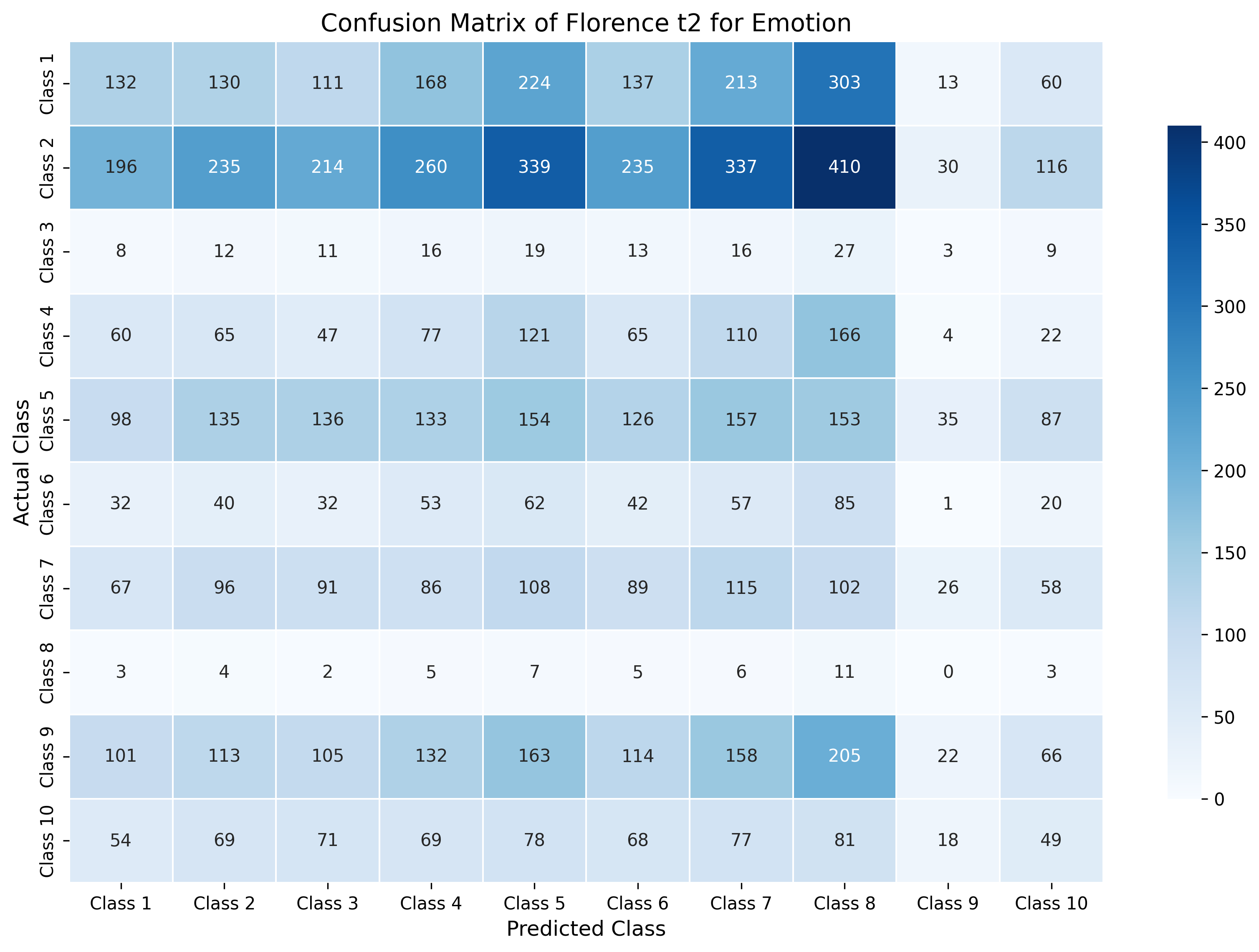}} \\[6pt]

\subfloat[LLaVa-1.5 (Action)\label{fig:LLaVaa}]
{\includegraphics[width=0.47\textwidth]{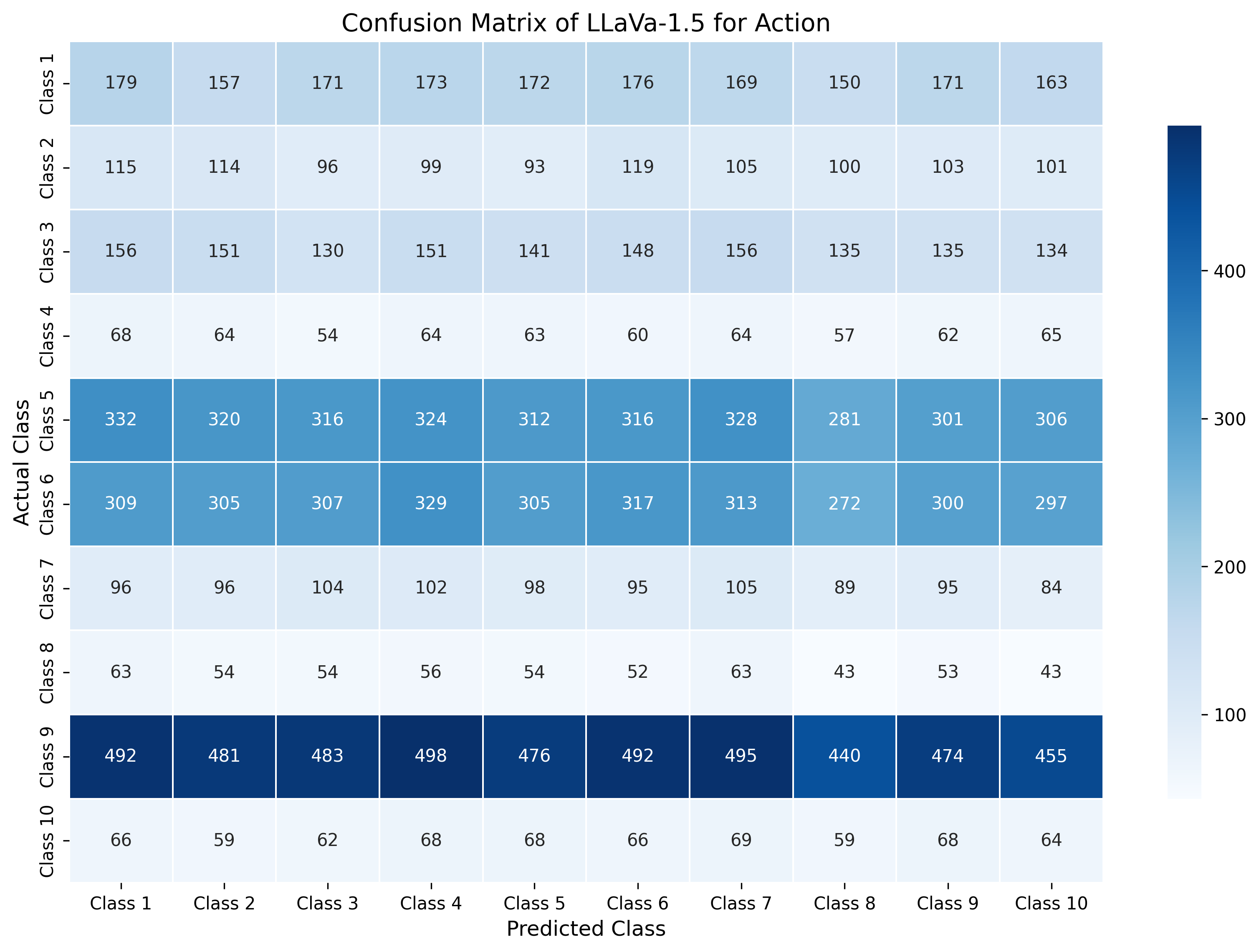}} &
\subfloat[LLaVa-1.5 (Emotion)\label{fig:LLaVae}]
{\includegraphics[width=0.47\textwidth]{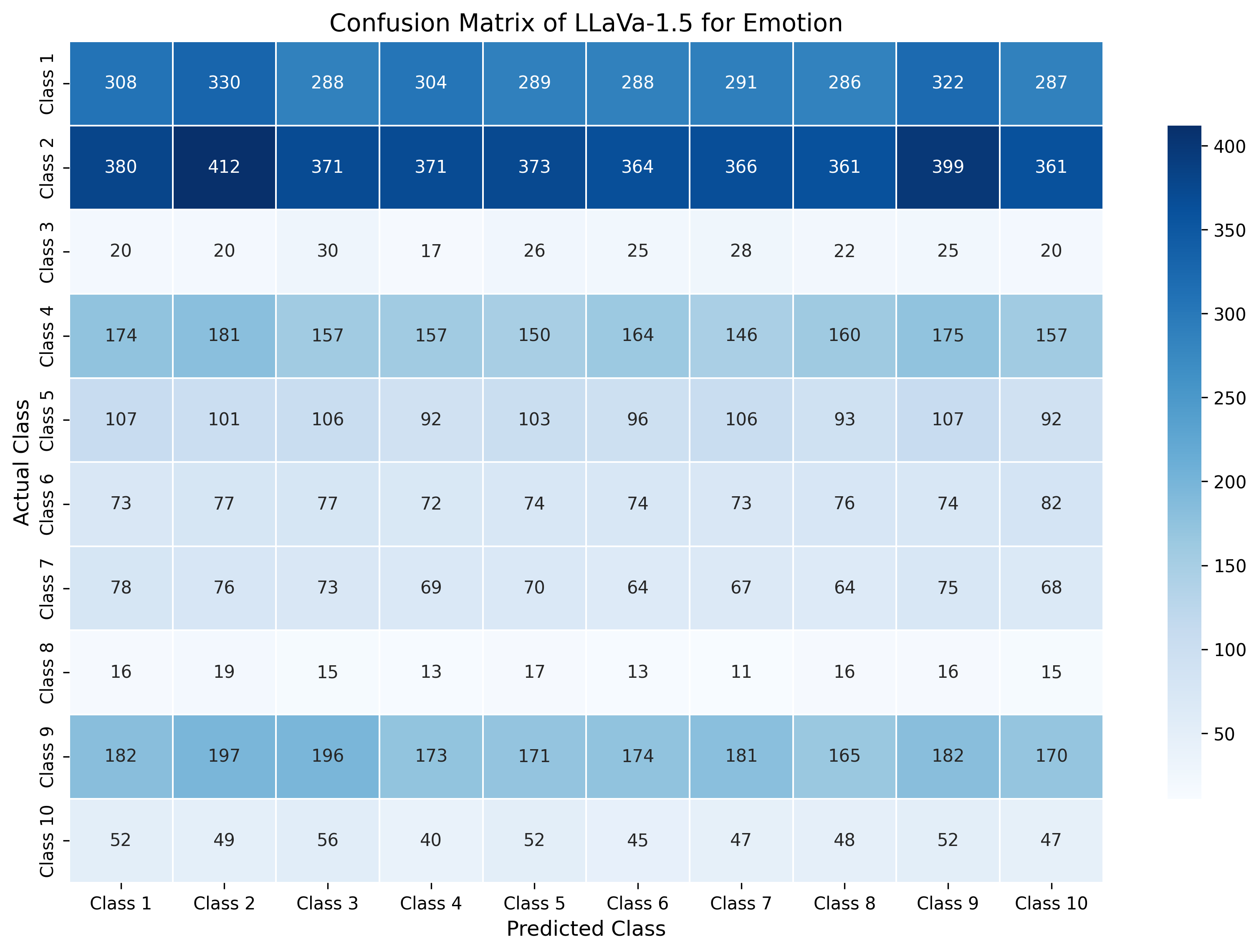}} \\
\end{tabular}

\caption{Confusion matrices for GPT-4o, Florence 2, and LLaVA-1.5 across action and emotion recognition tasks. \protect\\
\textbf{Action classes}: (1) Operating heavy machinery, (2) Lifting and carrying materials, (3) Measuring and marking surfaces, (4) Mixing cement or concrete, (5) Collaborating in teams, (6) Using hand tools (e.g., hammering, drilling), (7) Inspecting completed work, (8) Communicating with supervisors, (9) Following safety protocols, (10) Taking breaks/resting.\protect\\
\textbf{Emotion classes}: (1) Focused, (2) Determined, (3) Tired, (4) Alert, (5) Satisfied, (6) Anxious, (7) Proud, (8) Frustrated, (9) Cooperative, (10) Relieved.}

\label{fig:all_confusion}
\end{figure*}

In addition to aggregate metrics, we visualized the confusion matrices for each model and task to better understand the distribution of predictions across categories (Fig.~\ref{fig:all_confusion}). These visualizations reveal substantial misclassification trends, but also expose clear differences in each model’s capacity for class-specific resolution.

For the action recognition task, GPT-4o displayed a stronger concentration of predictions along the diagonal, particularly excelling at identifying ``Following safety protocols'' and ``Taking breaks/resting'' with dense and focused predictions. Despite this, some confusion remained between visually overlapping tasks such as ``Operating heavy machinery'' and ``Following safety protocols''. Florence 2’s confusion matrix appeared more diffusely populated, showing significant overlap across classes and indicating broader uncertainty in its predictions. LLaVa-1.5 showed the most dispersed pattern, lacking clear class-specific focus and frequently distributing predictions across semantically related activities, which corresponds with its lower class accuracies.

In the emotion recognition task, GPT-4o again demonstrated superior focus, especially in identifying ``Tired,'' ``Relieved,'' and ``Frustrated,'' with sharp peaks along the diagonal. However, confusion still arose between affectively similar emotions-such as ``Focused'' and ``Determined,''-highlighting the difficulty of parsing subtle emotional cues in static images. Florence 2 struggled more significantly in maintaining class separability, with notable misclassifications among the core emotional states. LLaVa-1.5, in particular, exhibited widespread dispersion across nearly all emotion classes, with little evidence of dominant correct predictions, suggesting a lack of effective internal representations for emotional categories.

These confusion matrices reinforce the challenges of fine-grained behavior and emotion recognition in real-world visual contexts. While GPT-4o shows promise in distinguishing both actions and emotional states, even it occasionally confuses classes with overlapping visual or contextual signals. These findings highlight the need for more contextual reasoning, temporal dynamics, and task-specific fine-tuning to enable robust interpretation of complex worker states on construction sites








\section{Discussion, Limitations, and Future Work}

This study examined how three general-purpose Vision-Language Models (VLMs)-GPT-4o, Florence 2, and LLaVa-1.5-performed in recognizing construction worker actions and emotional states from static imagery. Our goal was to understand off-the-shelf model performance, and in comparison to a human annotator. In summary, we find that while all models exhibited some level of competency, performance varied across models and categories, with several noteworthy trends and limitations. 

The static-image input format is a fundamental bottleneck for interpreting dynamic human behavior. Time-series data (e.g., video) or multimodal signals (e.g., audio, physiological sensors) are likely necessary to resolve ambiguities between lookalike categories. Our exploratory analysis can indicate that even advanced models benefit from domain-specific adaptation-whether through prompt tuning, curated data augmentation, or supervised fine-tuning on construction-specific datasets. 

GPT-4o consistently showed higher performance across metrics and confusion matrices, suggesting that its larger-scale training and multimodal alignment may support better generalization, even without domain-specific fine-tuning. Its strength in identifying visually distinct behaviors-such as “Taking breaks” or “Following safety protocols”-may reflect the model’s ability to recognize clear visual anchors. However, similar to the other models, GPT-4o still showed confusion between semantically close classes, which are difficult to distinguish without temporal or contextual cues.

Florence 2 and LLava-1.5 both showed strong patterns of misclassifications. LLaVa-1.5 exhibited the most frequent misclassifications and comparatively lower accuracy, particularly in emotion recognition. This can point to a less refined alignment between visual and linguistic modalities or limited pretraining exposure to emotionally nuanced imagery. While it captured some general features, its flat accuracy across classes suggests that improvements may require more targeted training or architectural adaptation.

Across all models, confusion matrices revealed that classes involving social interaction (e.g., collaborating in teams, and communication) and subtle emotional distinctions (e.g., “Anxious” vs. “Frustrated”) were more prone to error. This underscores the difficulty of interpreting human behavior from still images, where temporal dynamics and additional modalities (e.g., voice, motion, physiological cues) may be essential for disambiguation.

While this study provides valuable insights into the use of Vision-Language Models (VLMs) for detecting actions and emotions in construction site imagery, several limitations must be acknowledged. Recognizing these limitations provides essential context for interpreting the findings and highlights clear opportunities for future work.

First, although the dataset used in this work comprises 1000 images-substantially larger than earlier small-scale efforts-it may still not fully capture the breadth and variability of real-world construction environments. Certain actions and emotions, particularly rare, subtle, or context-dependent ones, may remain underrepresented. This uneven class distribution could introduce biases in model evaluation, potentially favoring more common behaviors and limiting the generalizability of the findings across different construction scenarios, environmental conditions, or cultural contexts.

Second, all annotations were derived from one-hot encoded labels based on static images, thereby omitting valuable temporal information that could enrich behavioral interpretation. Many construction-related actions and emotional expressions are inherently dynamic, evolving through sequences of motion, posture transitions, or facial changes. Restricting the analysis to single frames risks overlooking these temporal dependencies and may limit the models’ ability to detect fine-grained behavioral cues. 

Furthermore, the absence of multimodal data-such as audio signals, physiological measurements, or motion capture-constrains the models’ perceptual depth and hampers their capacity to differentiate complex worker states. Human annotation also introduces potential sources of error, including misinterpretation of subtle emotions or incorrect labeling. In some cases, an image may plausibly convey multiple simultaneous actions or emotional states, which are not captured by a one-hot encoding scheme.

Third, while GPT-4o, Florence 2, and LLaVa-1.5 represent state-of-the-art in general-purpose vision-language modeling, none were fine-tuned for construction-specific semantics, behaviors, or emotional cues. Their performance in this study reflects out-of-the-box capabilities and may not represent their true potential if subjected to domain-specific fine-tuning. Additionally, to maintain fairness across models, prompt engineering was deliberately kept minimal and consistent, but this conservative strategy may have limited the possibility of extracting optimal performance from models individually.

While we note that our work was an exploratory approach to current off-the-shelf models, a major part of our future work lies on expanding the dataset to include a larger and more diverse collection of annotated images-captured across a broader range of construction environments, lighting conditions, weather variations, and worker demographics-which enables more robust model training and evaluation. Incorporating time-series image sequences or video data could provide essential temporal context, improving the models’ ability to recognize both dynamic actions and evolving emotional states.

Integrating additional sensing modalities-such as audio signals, worker speech, body posture from skeleton tracking, or physiological data-could significantly enrich model inputs and enhance emotion recognition capabilities. These modalities would help disambiguate subtle affective states that are visually similar but contextually distinct, leading to more nuanced and reliable predictions.

Moreover, fine-tuning large models on construction-specific datasets or leveraging domain-adapted embeddings could improve classification precision and reduce confusion between overlapping action and emotion categories. Future work could also explore the development of lightweight or hybrid multimodal architectures optimized for real-time deployment on-site, such as wearable devices, mobile platforms, or smart surveillance systems.

Finally, calibrating and correcting model predictions can be performed through incorporating human-in-the-loop frameworks within such critical decision-making contexts. This approach would help ensure that VLM outputs remain interpretable, reliable, and actionable in high-stakes environments such as construction, where worker safety and operational efficiency are paramount.

\section{Conclusion}

This study explored the capability of general-purpose Vision-Language Models (VLMs) to recognize construction worker actions and emotional states from static imagery without domain-specific fine-tuning. Through systematic evaluation of GPT-4o, Florence 2, and LLaVa-1.5 on a curated dataset of 1000 construction site images, we found that while all models demonstrated some degree of generalization to this applied context, their performance varied substantially across tasks and categories. This work provides an initial benchmark and offers critical insights for future research at the intersection of computer vision, human-robot collaboration, and construction safety management using vision large language models.

\section{Acknowledgments}
The authors would like to thank the College of Engineering at Villanova University for their support through the Sports and Performance Engineering grant as well as the John Hennessy Undergraduate Research Fund.

\bibliographystyle{plain}  
\bibliography{sn-bibliography,nate-ref}

\phantomsection
\begin{IEEEbiography}[{\includegraphics[width=1in,height=1.25in,clip,keepaspectratio]{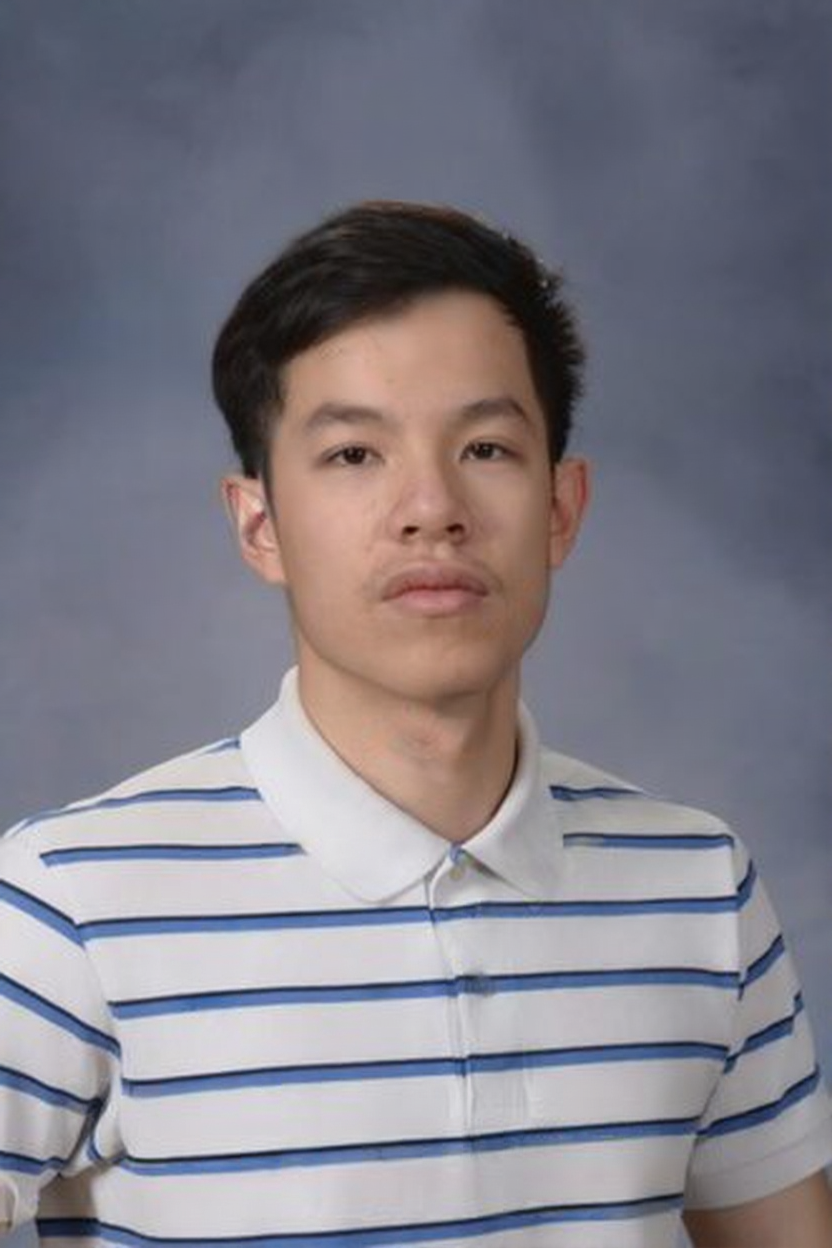}}]{Hieu T. Bui} Hieu Bui is an Undergraduate student at Villanova University in the department of Electrical and Computer Engineering. His research focuses on computer vision, specifically on image processing, autonomous driving and mapping.
\end{IEEEbiography}

\begin{IEEEbiography}[{\includegraphics[width=1in,height=1.25in,clip,keepaspectratio]{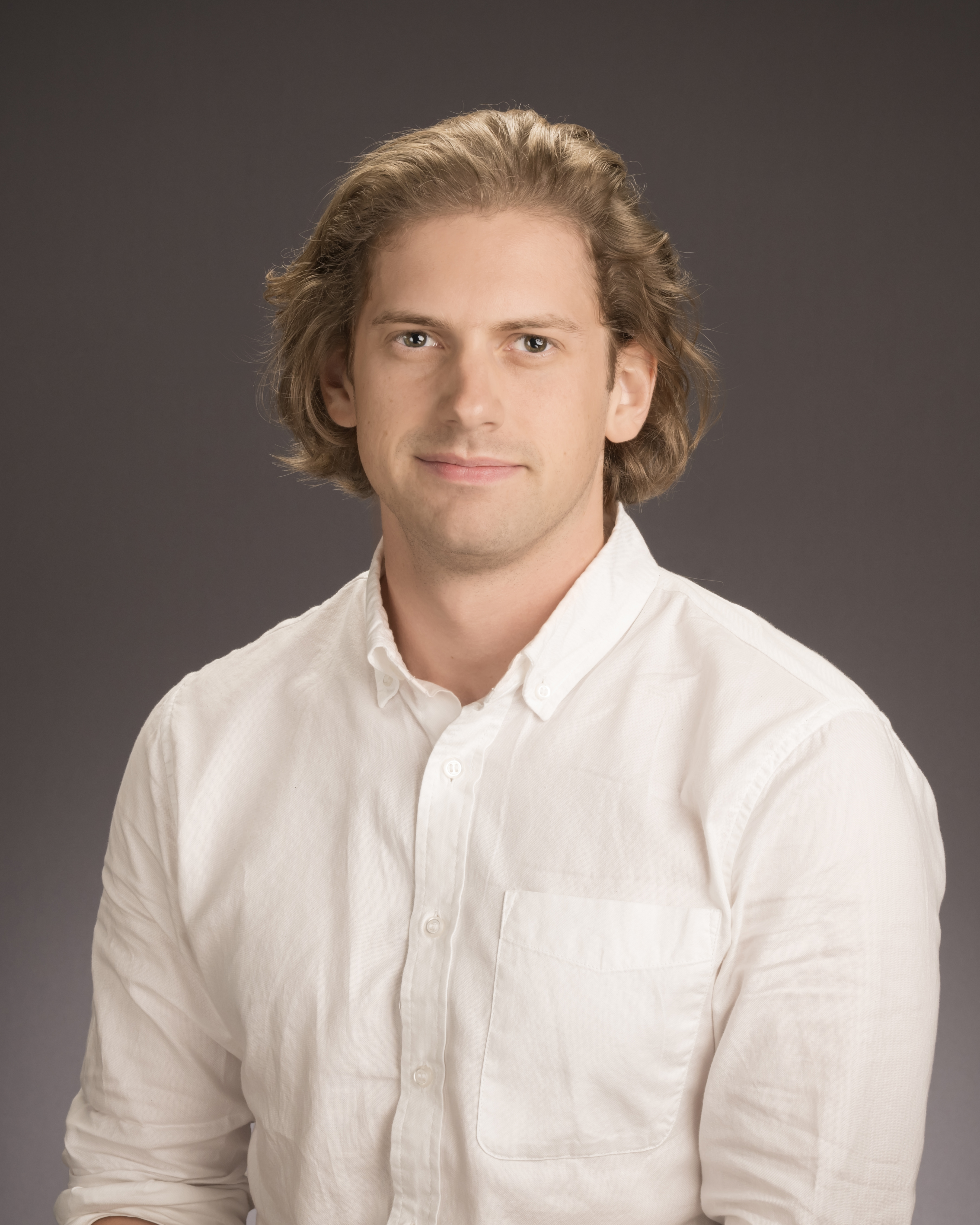}}]{Nathaniel E. Chodosh} Nathaniel E. Chodosh is an Assistant Professor in the Department of Computing Sciences at Villanova University. His research focuses on computer vision, and specifically on autonomous driving and LiDAR-based sensing. He received his PhD from the Robotics Institute at Carnegie Mellon University.
\end{IEEEbiography}

\begin{IEEEbiography}[{\includegraphics[width=1in,height=1.25in,clip,keepaspectratio]{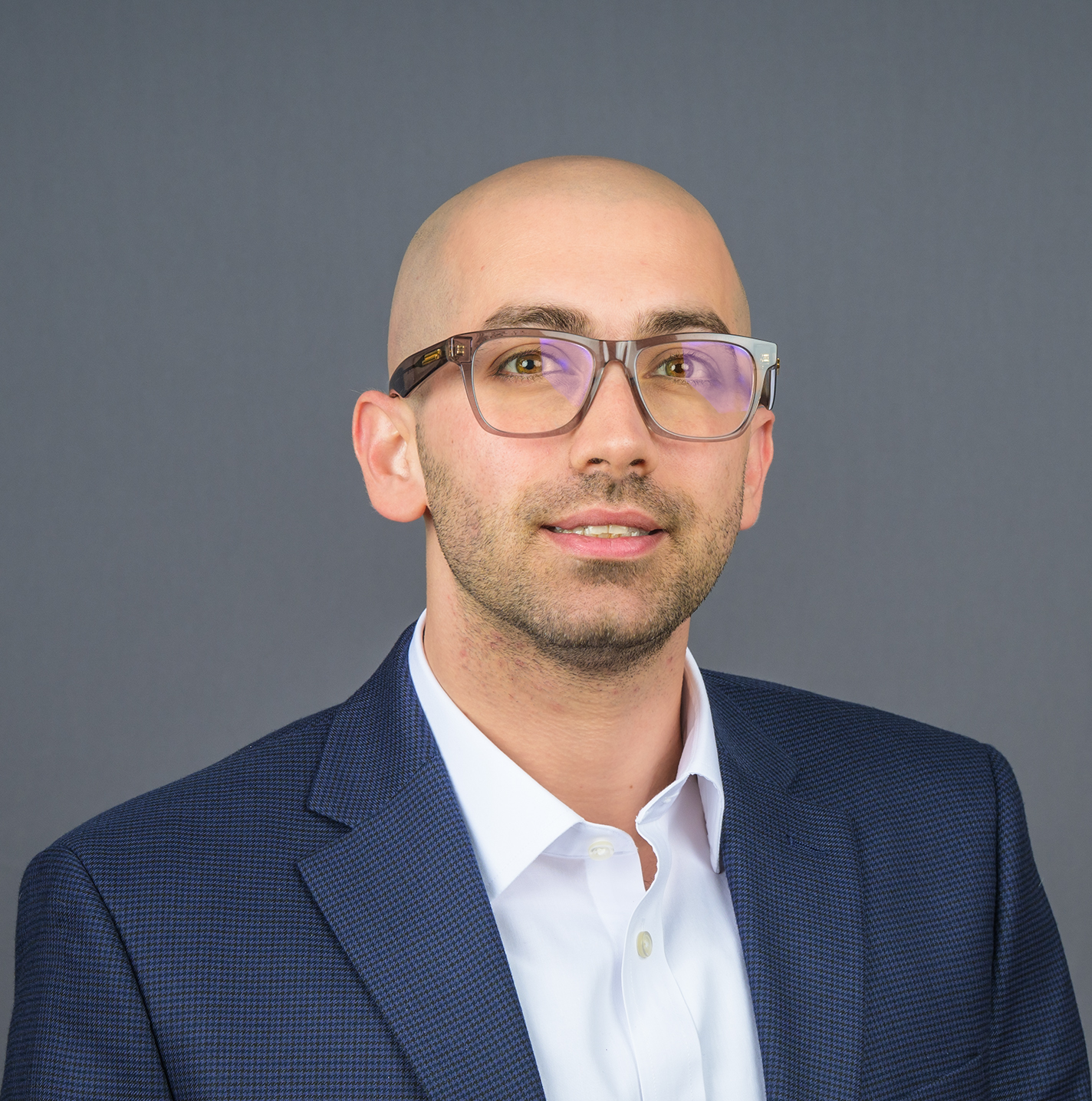}}]{Arash Tavakoli} Arash Tavakoli is an Assistant Professor of Transportation Engineering at Villanova University, where he directs the Human-Centered Cities Lab. His research focuses on developing civil and transportation systems that can understand, adapt to, and communicate with humans. Before joining Villanova, he was a Postdoctoral Scholar at Stanford University. He earned his Ph.D. in Civil Engineering from the University of Virginia.
\end{IEEEbiography}

\end{document}